\documentclass[twocolumn]{article}
\usepackage[top=2cm, bottom=2.5cm, left=2cm, right=2cm]{geometry}
\usepackage{times,textcomp}
\usepackage[T1]{fontenc}
\usepackage[utf8]{inputenc}
\usepackage{amssymb,amsmath}
\usepackage{enumitem}
\usepackage{eurosym}
\usepackage{listings}
\usepackage{float,caption}
\usepackage{diagbox}
\usepackage{tabularx}

\usepackage[giveninits=true]{biblatex}
\setlist[itemize]{itemsep=0pt}

\usepackage{titlesec} 
\titleformat{\section}{\large\bfseries}{\thesection}{1em}{\MakeUppercase} 
\titleformat{\subsection}{\large\bfseries}{\thesubsection}{1em}{}
\titlespacing*{\section}{0pt}{\parskip}{0pt}
\titlespacing*{\subsection}{0pt}{\parskip}{0pt}

\usepackage{longtable,booktabs}
\usepackage{graphicx,grffile}
\usepackage{fancyhdr}

\title{Bayesian and Dempster-Shafer models for combining multiple sources of evidence in a fraud detection system}

\usepackage{authblk}

\author{Fabrice Daniel}

\affil{\small Artificial Intelligence Department of Lusis, Paris, France\\http://www.lusisai.com}

\date{March 2021}
\begin{document}

\maketitle

\begin{abstract}
Combining evidence from different sources can be achieved with Bayesian or Dempster-Shafer methods. The first requires an estimate of the priors and likelihoods while the second only needs an estimate of the posterior probabilities and enables reasoning with uncertain information due to imprecision of the sources and with the degree of conflict between them. This paper describes the two methods and how they can be applied to the estimation of a global score in the context of fraud detection.
\end{abstract}


\section{Introduction}

Fraud detection mainly relies on expert driven methods that implement a set of rules and data driven approaches implementing machine learning (ML) models. Both provide an estimate (or a score) for a new transaction to be fraudulent.

While each ML model naturally returns a fraud probability, the experts can also attach a probability to each rule. They can also be automatically calculated from the labeled history. Combining them together produces a global score that can be used in a near real time system to rank a set of transactions having the highest probability to be fraudulent. By obtaining this ranking, investigators can concentrate their efforts on the suspect transactions with the highest probability of being true frauds. 

The most common approaches for combining scores are summing individual scores or returning the highest score among the trigged rules. This is not entirely satisfactory given that summing scores is equivalent to averaging the probabilities returned by each predictor (rule or model). It also does not take into account the uncertainty of each predictor and the degree of conflict between them.

For the Lusis fraud system, we work on implementing more appropriate approaches.

This paper proposes two ways for addressing this problem. The first is to use \textbf{Bayesian} methods \cite{mosoCreditCardFraud2018}; the second is to combine the scores by using \textbf{Dempster-Shafer} theory \cite{panigrahiCreditCardFraud2009}.

\section{Bayesian approach}

\subsection{Fundamental concepts}

\begin{itemize}
	\item Let $E_i$ be the evidence $i$ that corresponds to a fraud detection rule (ML model) with $1 \leq i \leq n$,
	\item Let $E = \{E_1, \ldots, E_n\}$ be a set of evidence,
	\item There are two hypotheses, $H_f$ and $H_g$, corresponding to a fraudulent and genuine transaction respectively,
	\item Let $P(H_f|E_i)$ be the probability of transaction being fraudulent given $E_i$,
	\item Let $P(H_g|E_i)$ be the probability of transaction being genuine given $E_i$,
	\item Let $P(E_i|H_f)$ be the probability of $E_i$ being triggered given $H_f$, and is called the likelihood.
\end{itemize}

In the context of a Fraud Detection System (FDS), $P(H_f|E_i)$ corresponds to the output of a ML model $E_i$ or to the probability attached by an expert to rule $E_i$.
\begin{align}
	P(H_f|E) &= \frac{P(E|H_f)P(H_f)}{P(E)} \nonumber \\
			 &= \frac{P(E_1,\dots,E_n|H_f)P(H_f)}{P(E_1,\dots,E_n)} \label{eq:PE}
\end{align}
Because $P(E_1,\dots,E_n|H_f)$ in \eqref{eq:PE} is intractable we assume that the events $E_i$ are independent, so we can use a naive Bayes model to compute the combined probability $P(H_f|E)$. 
\begin{align}
P(H_f|E) &= \frac{P(E_1|H_f)P(E_2|H_f)...P(E_n|H_f)P(H_f)}{P(E_1,\dots,E_n)}  \nonumber \\
          &= P(H_f)\frac{\prod_{i=1}^{n} P(E_i|Hf)}{P(E_1,\dots,E_n)}
\end{align}
And because
\begin{equation}
P(E) = P(H_f)P(E|H_f) + P(H_g)P(E|H_g)
\end{equation}
we find 
\begin{subequations}
\begin{align}
	P(H_f|E) &= \frac{1}{Z}P(H_f)\prod_{i=1}^{n}P(E_i|H_f) \\
	P(H_g|E) &= \frac{1}{Z}P(H_g)\prod_{i=1}^{n}P(E_i|H_g)		
\end{align}
\end{subequations}
Where Z is a normalization factor:
\begin{align}
	Z &= P(E) \nonumber \\
	  &= P(H_f)P(E|H_f) + P(H_g)P(E|H_g) \nonumber \\
	 &= P(H_f)\prod_{i=1}^{n} P(E_i|Hf) + P(H_g)\prod_{i=1}^{n} P(E_i|Hg)
\end{align}
When $n$ becomes large, there is a risk of vanishing precision. To fix this issue we can apply a logarithm to transform the product into a sum.

The main shortcoming of this approach is the independence assumption that is not the case in most of the real problems. In \cite{zhangOptimalityNaiveBayes2004} an analysis of the Bayesian classification problem showed that there are sound theoretical reasons for the surprising implausible efficacy of naive Bayes classifiers.

\subsection{Example}

Table \ref{table:bayes} shows an example with numerical values from \cite{mosoCreditCardFraud2018}.

Let $E_1$ and $E_2$ be triggered rules or ML model outputs.

The dataset has 30 transactions including 7 frauds.

\begin{table}[h!]
	\centering
	\begin{tabular}{|l|l|l|l|l|}
	\hline
	 & fraud & genuine & $P(E_i|H_f)$ & $P(E_i|H_g)$ \\ 
	\hline
	$E_1$ & 4 & 6 & 0.57 & 0.26 \\ 
	\hline
	$E_2$ & 1 & 2 & 0.14 & 0.09 \\ 
	\hline
	\end{tabular}
\caption{Example with two pieces of evidence \(E_1\) and \(E_2\)}
\label{table:bayes}
\end{table}

This table means that:

\begin{itemize}
	\item $E_1$ is triggered by 10 transactions out of 30 in the dataset
	\item When $E_1$ is observed, 4 transactions were true frauds, and 6 were not
	\item $P(E_1|H_f)$ is the likelihood, the probability of observing $E_1$, given $H_f$, meaning that a transaction is fraudulent. Among the 7 frauds observed, 4 of them trigger the rule $E_1$
	\item $P(E_1|H_g)$ is the likelihood, the probability of observing $E_1$, given $H_g$, meaning that a transaction is genuine. Among the 23 genuine transactions observed, 6 of them trigger the rule $E_1$
\end{itemize}

First we estimate prior probabilities:
\begin{align*}
P(H_f) &= \frac{7}{30} = 0.23 \\
P(H_g) &= \frac{23}{30} = 0.77
\end{align*}
Assume that we observe the set $E = \{E_1, E_2\}$ of evidence.
Then let's compute the likelihoods:
\begin{align*}
P(E|H_f) &= \prod_{i=1}^{n} P(E_i|H_f) \\
		 &= P(E_1|H_f).P(E_2|H_f) \\
		 &= 0.57 \times 0.14 \\
		 &= 0.0798 \\\\
P(E|H_g) &= \prod_{i=1}^{n} P(E_i|H_g) \\
		 &= P(E_1|H_g).P(E_2|H_g) \\
		 &= 0.26 \times 0.09 \\
		 &= 0.0234
\end{align*}
This means the likelihood of this transaction being a fraud is 0.0798.

We then compute the marginal likelihood $P(E)$:
\begin{align*}
P(E) &= P(H_f)P(E|H_f) + P(H_g)P(E|H_g) \\
	 &= 0.23 \times 0.0798 + 0.77 \times 0.0234 \\
	 &= 0.0184 + 0.0180 \\
	 &= 0.0364
\end{align*}
So the posterior probabilities are:
\begin{align*}
P(H_f|E) &= \frac{P(E|H_f).P(H_f)}{P(E)} \\
	 &= \frac{0.0798 \times 0.23}{0.0364} \\
	 &= 0.504 \\\\
P(H_g|E) &= \frac{P(E|H_g).P(H_g)}{P(E)} \\
	 &= \frac{0.0234 \times 0.773}{0.0364} \\
	 &= 0.495
\end{align*}
The probability of this transaction being a fraud is 0.504.

\section{Dempster-Shafer approach}

Using Naive Bayes is only possible if we can obtain priors and likelihoods estimates from an expert or from historical data.

In most cases we only have the posterior probabilities attached to rules or models. It can comes from an \textbf{expert estimate}\footnotemark attached to each rule or from a machine learning model probability prediction.

\footnotetext{When an expert attaches 0.75 to a rule, he estimates that when this rule is triggered it has 75\% chances to be a fraud.}

A way to combine probabilities of fraudulence given individual rules is to use Dempster-Shafer theory.

Dempster-Shafer theory (DST) provides a framework for combining different sources of evidence into a global belief for a given hypothesis \cite[36]{bellengerSemanticDecisionSupport2013}.

\subsection{Fundamental concepts}

Let $\Omega$ be the universe of all the possible states, meaning the set of all the $N$ hypothesis, also called the \textit{frame of discernment}.
\begin{equation}
	\Omega=\{H_1,\dots,H_N\}
\end{equation}
We can define a set $2^\Omega$, named the \textit{power set}, that contains all the possible subsets of $\Omega$, including the empty set.
\begin{equation}
	2^\Omega : \{\emptyset, \{H_1\}, \dots, \{H_N\}, \{H_1,H_2\},\dots,\dots,\Omega\}
\end{equation}
In our case we assume a universe of two hypotheses $H_f$ and $H_g$ for fraudulent and genuine transactions.
\begin{equation}
	2^\Omega : \{\emptyset, \{H_f\},\{H_g\}, \{H_f,H_g\}\} \label{eq:2omega}
\end{equation}
The theory of evidence assigns a belief mass to each element of the \textit{power set}. Formally the mass function, called the \textit{basic mass assignment} (BMA), \textit{basic belief assignment} (BBA) or \textit{basic probability assignment} (BPA) depending on the source\footnotemark, is defined by:
\footnotetext{basic probability assignment (BPA) seems to be the most frequent naming}
\begin{equation}
	m: 2^\Omega \rightarrow [0,1]
\end{equation}
First, the mass of the empty set is zero:
\begin{equation}
	m(\emptyset)=0
\end{equation}
Second, the masses of all the members of the power set add up to a total of 1:
\begin{equation}
	\sum_{A\in 2^\Omega}m(A) = 1 \label{eq:mA}
\end{equation}
The mass $m(A)$ in \eqref{eq:mA} is interpreted as the part of belief placed strictly on A. It expresses the proportion of all relevant and available evidence that supports the claim that the actual state belongs to A but to no particular subset of A. This quantity differs from a probability since the total mass can be given either to singleton hypotheses $H_n$ or to composite ones $A$ \cite[38]{bellengerSemanticDecisionSupport2013}.

While a probability can only be assigned to the singletons $H_f$ or $H_g$, a mass $m(A)$ can also be assigned to the composition $\{H_f,H_g\}$.

Belief mass on $m(A)$ where A is a singleton is interpreted as: \textit{$A$ is true}.

Belief mass on $m(A)$ where A is non-atomic is interpreted as: \textit{one of the $A$ components is true, but the source is uncertain about which one of them is true}.

Elements of $\Omega$ having $m(A) \neq 0$ are called \textit{focal elements}.

\subsection{Belief, Plausibility, uncertainty and probability interval}

The belief (also named credibility) $bel(A)$ for a set $A$ is defined as the sum of all the masses of subsets of the set of interest:
\begin{equation}
	bel(A) = \sum_{B|B \subseteq A}m(B) \quad \forall A \subseteq \Omega
\end{equation}
So in our case where $A \in 2^\Omega$ and $2^\Omega$ defined in \eqref{eq:2omega}, we find
\begin{subequations}
\begin{align}
	bel(H_f) &= m(H_f) \label{eq:bel_a}\\
	bel(H_g) &= m(H_g) \label{eq:bel_b}
\end{align}
\end{subequations}
The plausibility $pl(A)$ is the sum of all the masses of the sets $B$ that intersect the set of interest $A$:
\begin{equation}
	pl(A) = \sum_{B|B \cap A \neq \emptyset}m(B) \quad \forall A \subseteq \Omega
\end{equation}
So in our case:
\begin{subequations}
\begin{align}
	pl(H_f) &= m(H_f) + m(H_f,H_g) \\
	pl(H_g) &= m(H_g) + m(H_f,H_g)
\end{align}
\end{subequations}
It can be shown that
\begin{equation}
	bel(A) \leq pl(A) \quad \forall A \subset \Omega
\end{equation}
This equation can be interpreted as "certain implies plausible" \cite[40]{bellengerSemanticDecisionSupport2013}.

Plausibility and belief are related to each other as follows:
\begin{equation}
	pl(A) = 1 - bel(\bar{A}) \quad \forall A \subset \Omega
\end{equation}
According to \cite[41]{bellengerSemanticDecisionSupport2013} a probability interval can be defined as the interval with $bel(A)$ and $pl(A)$ as its lower and upper bound respectively
\begin{equation}
	bel(A) \leq P(A) \leq pl(A)
\end{equation}
The difference between $pl(A)$ and $bel(A)$ is the ignorance about a specific hypothesis $A$. 

The author also states that :

\begin{quote}
	\emph{``if focal sets are only singletons (i.e.~we assign only masses to singleton hypothesis), then the mass distributions, credibility measures, plausibility ones and commonalities are merged and coincide with a probability distribution.''}
\end{quote}

In our case, we are only assigning a mass to each of the triggered Fraud detection rules. For each of them, an expert has attached a score, meaning a probability to be a fraud given that it has been triggered. Each rule or machine learning model only returns the probability for a transaction to be a fraud. So the focal set are only singletons, meaning we only have the following masses defined : $m_i(H_f)$ and $m_i(H_g)$, where $i$ is the $i^{th}$ triggered rule.
\begin{subequations}
\begin{align}
	m_i(\bar{H_f}) &= m_i(H_g) \\
	m_i(\bar{H_g}) &= m_i(H_f) \\
	m_i(H_f,H_g)&=0
\end{align}
\end{subequations}
This means in our case:
\begin{equation}
	P(A) = bel(A) = pl(A) \label{eq:PA}
\end{equation}
If we want to also consider the rules not triggered in the model, we should assign the whole mass to the uncertainty with $m_j(H_f,H_g)=1$ where $j$ is the $j^{th}$ not-triggered rule.

Here, we only consider the masses on the triggered rules.

%--------------------------------------
% Dempster’s rule of combination
%--------------------------------------
\subsection{Dempster’s rule of combination}

When several rules are triggered, we want to calculate the probability for a transaction to be a fraud. 

Dempster-Shafer proposes a combination rule for calculating the set of masses $m_{1,2}$ from $m_1$ and $m_2$.
\begin{align}
	m_{1,2}(A) &= (m_1 \oplus m_2)(A) \nonumber \\
           	   &= \frac{1}{1-K}\sum_{B \cap C=A \neq \emptyset}m_1(B)m_2(C) \label{eq:m12}
\end{align}
Where $K$ is a measure of the degree of conflict between two mass sets. $1-K$ is the \textit{normalization factor}. 
\begin{align}
	K &= \sum_{B \cap C=\emptyset}m_1(B)m_2(C) \nonumber \\
  	  &=(m_1 \oplus m_2)(\emptyset) \quad K \in [0,1] \label{eq:K}
\end{align}
Having $K$ near to $0$ means there is small conflict between the two mass sets and $1$ means that they are in total conflict.

From \eqref{eq:m12} and \eqref{eq:K} we can compute $m_{1,2}(H_f)$ and $m_{1,2}(H_g)$:
\begin{subequations}
\begin{align}
	K &= (m_1(H_f).m_2(H_g)) \nonumber \\
	&+(m_1(H_g).m_2(H_f)) \\
	m_{1,2}(H_f) &= \frac{m_1(H_f).m_2(H_f)}{1-K} \\
	m_{1,2}(H_g) &= \frac{m_1(H_g).m_2(H_g)}{1-K}
\end{align}
\end{subequations}
After getting the combined mass $m_{1,2}$ we need to get the corresponding beliefs. In our case we saw with \eqref{eq:bel_a} and \eqref{eq:bel_b} that they can be directly derived from $m$, so:
\begin{subequations}
	\begin{align}
		bel(H_f) &= m_{1,2}(H_f) \\
		bel(H_g) &= m_{1,2}(H_g)
	\end{align}
\end{subequations}
And as seen with \eqref{eq:PA}: $P(A) = bel(A)$

So we get:
\begin{subequations}
\begin{align}
	P(H_f) &= bel(H_f) \\
	P(H_g) &= bel(H_g)
\end{align}
\end{subequations}
%
%--------------------------------------
% Combining more than two sources
%--------------------------------------
\subsection{Combining more than two sources}

Equation \eqref{eq:m12} is relative to two masses sets only. In a rule engine or more generally in a multiple source decision system (rules + several machine learning models) we need to potentially handle many rules that can be triggered together on the same transaction.

So we need to be able to compute : 
\begin{equation}
	(m_1 \oplus m_2 \oplus \dots \oplus m_n)(A)
\end{equation}
According to \cite{dezertMathematicalTheoryEvidence2011}
\begin{align}
	m_1 \oplus m_2 \oplus m_3 &= (m_1 \oplus m_2) \oplus m_3 \nonumber \\
							  &= m_1 \oplus (m_2 \oplus m_3) \nonumber \\
							  &= m_2\oplus (m1 \oplus m_3)
\end{align}
We use this property to combine as many sources as needed.

Now we have a framework to compute the probability of a fraud from several sources of evidence.

%--------------------------------------
% Examples
%--------------------------------------
\subsection{Examples}

%--------------------------------------
% Without uncertainty
%--------------------------------------
\subsubsection{Without uncertainty}

Table \ref{table:dempster_no_uncertainty} shows an example with numerical values from \cite{chenDataClassificationUsing2014}.

Assume two sources $S_1$ and $S_2$ (rules or models) providing sets of masses $m_1$ and $m_2$ respectively.

\begin{table}[h!]
	\centering
	\begin{tabular}{|l|l|l|}
		\hline
		\diagbox[width=2.7cm]{\(S_2\)}{\(S_1\)} & \(m_1(H_f)=0.6\) & \(m_1(H_g)=0.4\) \\ 
		\hline
		\(m_2(H_f)=0.8\) & 0.48 & 0.32 (conflict) \\ 
		\hline
		\(m_2(H_g)=0.2\) & 0.12 (conflict) & 0.08 \\ 
		\hline
	\end{tabular}
	\caption{Two sources and no mass assignment to uncertainty}
	\label{table:dempster_no_uncertainty}
\end{table}

Then, as per the Dempster-Shafer combination rule:
\begin{align*}
K &= (m_1(H_f).m_2(H_g)) \\ 
  &+ (m_1(H_g).m_2(H_f)) \\
  &= (0.6 \times 0.2) + (0.4 \times 0.8) \\
  &= 0.44 \\
(m_1 \oplus m_2)(H_f) &= \frac{1}{1-0.44}m_1(H_f).m_2(H_f) \\
                      &= \frac{0.48}{0.56} \\
					  &= 0.8571 \\
(m_1 \oplus m_2)(H_g) &= \frac{1}{1-0.44}m_1(H_g).m_2(H_g) \\
                      &= \frac{0.08}{0.56} \\
					  &= 0.1428
\end{align*}
Since, in our case, focal sets are singletons:
\begin{equation*}
	P(H_f) = bel(H_f) = m_{1,2}(H_f) = 0.8571
\end{equation*}
The probability of this transaction being a fraud is 0.8571.

%--------------------------------------
% With uncertainty
%--------------------------------------
\subsubsection{With uncertainty}

A major advantage of Dempster-Shafer is its capacity to consider uncertainty. It gives the ability to return a probability interval instead of a point estimate. 

To illustrate this principle, let us assume the fraud detection rule engine allows the experts to assign an uncertainty mass to any rule.

In table \ref{table:dempster_with_uncertainty} we assume that the expert assigns masses $m_1(H_f,H_g) = 0.2$ and $m_2(H_f,H_g) = 0.5$ to quantify respectively the uncertainty of the sources $S_1$ and $S_2$. 

\begin{table*}[t]
	\centering
	\begin{tabular}{|l|l|l|l|}
		\hline
		\diagbox[width=3.5cm]{\(S_2\)}{\(S_1\)} & \(m_1(H_f)=0.7\) & \(m_1(H_g)=0.1\) &
		\(m_1(H_f,H_g)=0.2\) \\ 
		\hline
		\(m_2(H_f)=0.3\) & 0.21 & 0.03 (conflict) & 0.06 \\ 
		\hline
		\(m_2(H_g)=0.2\) & 0.14 (conflict) & 0.02 & 0.04 \\ 
		\hline
		\(m_2(H_f,H_g)=0.5\) & 0.35 & 0.05 & 0.10 \\ 
		\hline
	\end{tabular}
	\caption{Two sources and masses assignment to uncertainty}
	\label{table:dempster_with_uncertainty}
\end{table*}

\begin{table*}[t]
	\centering
	\begin{tabular}{|l|l|l|l|}
		\hline
		\diagbox[width=3.5cm]{\(S_2\)}{\(S_1\)} & \(m_1(H_f)=0.7\) & \(m_1(H_g)=0.2\) &
		\(m_1(H_f,H_g)=0.1\) \\ 
		\hline
		\(m_2(H_f)=0.3\) & 0.21 & 0.06 (conflict) & 0.03 \\ 
		\hline
		\(m_2(H_g)=0.6\) & 0.42 (conflict) & 0.12 & 0.06 \\ 
		\hline
		\(m_2(H_f,H_g)=0.1\) & 0.07 & 0.02 & 0.01 \\ 
		\hline
	\end{tabular}
	\caption{Two sources and reduced masses assignment to uncertainty}
	\label{table:dempster_reduced_uncertainty}
\end{table*}

So we compute :
\begin{align*}
	K &= (m_1(H_f).m_2(H_g)) \\
	  &+ (m_1(H_g).m_2(H_f)) \\
	  &= (0.6 \times 0.2) + (0.4 \times 0.3) \\
	  &= 0.17 \\
	(m_1 \oplus m_2)(H_f) &= \frac{1}{1-0.17}m_1(H_f).m_2(H_f) \\
	                      &= \frac{0.21}{0.83} \\
						  &= 0.253 \\
	(m_1 \oplus m_2)(H_g) &= \frac{1}{1-0.24}m_1(H_g).m_2(H_g) \\
	                      &= \frac{0.08}{0.83} \\
						  &= 0.024
\end{align*}
\begin{align*}
	(m_1 \oplus m_2)(H_f,H_g) &= \frac{1}{1-0.24}(m_1(H_f).m_2(H_f,H_g) \\
	&+ m_1(H_g).m_2(H_f,H_g) \\ 
	&+ m_2(H_f).m_2(H_f,H_g) \\ 
	&+ m_2(H_g).m_2(H_f,H_g) \\
	&+ m_1(H_f,H_g).m_2(H_f,H_g))\\
	                      &= \frac{0.6}{0.83} \\
						  &= 0.723 \\\\
	pl(H_f) &= m(H_f) + m(H_f,H_g) \\
	&= 0.253 + 0.723 \\
	&= 0.976
\end{align*}
The true probability for this transaction to be a fraud is in this interval:
\begin{align*}
	bel(Hf) &\leq P(H_f) \leq pl(H_f)\\
	0.25 &\leq P(H_f) \leq 0.98
\end{align*}
Now let's study, with the example from table \ref{table:dempster_reduced_uncertainty}, what happens if after a while, the expert gets more statistics about these rules, such that they can reduce the uncertainty.

In this case we find that the probability interval is reduced, with the minimum probability increasing from $0.25$ to $0.40$ while the masses $m_1(H_f)$ and $m_2(H_f)$ remain unchanged.
\begin{equation*}
	0.40 \leq P(H_f) \leq 0.77
\end{equation*}
With this information the fraud scoring engine is able to provide a more informative ranking of the transactions, by including an estimate of the uncertainty. 

Some transactions, not detected as fraudulent with a point estimate, become suspect when uncertainty is taken into consideration.

For instance if we modify the previous example by removing uncertainty, allocating the remaining mass to $H_g$ such as $m_1(H_g) = 0.3$ and $m_2(H_g)=0.7$, we find $P(H_f) = 0.5$; if the threshold for a fraud detection is set to $\tau > 0.5$ then this transaction is not considered fraudulent, while in the previous case it is considered suspicious.

Now if we reduce the uncertainty mass by half, and distribute the remaining mass equally between $H_f$ and $H_g$ by adding $0.025$ to each of them, then we find
\begin{equation*}
	0.50 \leq P(H_f) \leq 0.70
\end{equation*}
The interval is reduced and the credibility increases up to $0.5$, so the ranking algorithm can decide to give it a better rank even though the plausibility is lower ($0.70$ compared to $0.77$). 

\section{Conclusion}

Combining probability estimates of fraud detection rules and ML models predictions by using Dempster-Shafer has two advantages compared to a pure Bayesian approach. First it's applicable to any situation where knowledges or historical data are not available to estimate the prior probabilities. Second it can represents the level of uncertainty, providing an interval instead of only a point estimate for the true probability.

A future improvement study could be focused on how to weight ML models or some specific rules. Such weighting can be useful to reflect the business impact in term of cost\footnotemark for a given rule or model.

Another future study could be focused on ranking the combined probability estimates of suspect transactions not only from the point estimates but also by using their respective uncertainty.

\footnotetext{the cost concept depends on the context, it can be monetary but it can also be reputation, time or anything else. According to literature such as \cite{freryEnsembleLearningExtremely2019} and \cite{pozzoloAdaptiveMachineLearning2015} this concept is known to be very hard to quantify in the context of payments fraud}

\newpage

\printbibliography

\end{document}